\documentclass[10pt,letterpaper]{article}
\usepackage[T1]{fontenc}
\usepackage{spconf,amsmath,amssymb,graphicx,booktabs,array}
\usepackage[hidelinks]{hyperref}
\hypersetup{pdftitle={TIMBRE: Teaching Time Series Forecasters to Read, Remember, and Reconcile},pdfsubject={Temporal Integration of Memory-Based Responses and Evidence},pdfkeywords={time series forecasting, TIMBRE, event memory, evidence fusion}}
\DeclareMathSizes{9}{10}{7}{5}

\newcommand{\Hh}{\mathbf{H}}
\newcommand{\calL}{\mathcal{L}}
\newcommand{\projectlink}{\url{https://github.com/stephen-guan-researcher/TIMBRE}}
\newcommand{\weightlink}{\url{https://huggingface.co/XinyuGuan/TIMBRE}}
\title{\fontsize{14}{16}\selectfont TIMBRE: Teaching Time Series Forecasters\\to Read, Remember, and Reconcile}
\hypersetup{pdfauthor={Xinyu Guan, Zhirong Zhang, Hongyuan Liu, Pengcheng Xu, Yu Sun, Chen Song, Qianyang Zhao}}
\name{\begin{tabular}{@{}c@{}}
\href{https://orcid.org/0009-0006-8121-3308}{Xinyu Guan}\textsuperscript{1,*,$\dagger$},
\href{https://orcid.org/0009-0009-9734-7126}{Zhirong Zhang}\textsuperscript{2,*},
\href{https://orcid.org/0009-0001-7692-6518}{Hongyuan Liu}\textsuperscript{3,4},
\href{https://orcid.org/0009-0000-9858-9316}{Pengcheng Xu}\textsuperscript{5},\\
\href{https://orcid.org/0009-0009-4419-7611}{Yu Sun}\textsuperscript{6},
\href{https://orcid.org/0009-0003-7972-5353}{Chen Song}\textsuperscript{7},
Qianyang Zhao\textsuperscript{1}
\end{tabular}}
\address{\begin{tabular}{@{}c@{}}
\normalsize\textsuperscript{1}Alibaba Group\quad\textsuperscript{2}Chang'an University\quad\textsuperscript{7}Xiaomi\\
\normalsize\textsuperscript{3}University of Electronic Science and Technology of China\quad\textsuperscript{4}University of Bristol\\
\normalsize\textsuperscript{5}Computer Science, University of California, Irvine\quad\textsuperscript{6}University of Pennsylvania\\
\normalsize\textsuperscript{1}\href{mailto:guanhan.gxy@alibaba-inc.com}{guanhan.gxy@alibaba-inc.com}\quad
\textsuperscript{1}\href{mailto:zhaoqianyang.zqy@alibaba-inc.com}{zhaoqianyang.zqy@alibaba-inc.com}\quad
\textsuperscript{2}\href{mailto:zhangzhirong@chd.edu.cn}{zhangzhirong@chd.edu.cn}\\
\normalsize\textsuperscript{3}\href{mailto:herkliu1@gmail.com}{herkliu1@gmail.com}\quad
\textsuperscript{5}\href{mailto:pengcheng.xu@uci.edu}{pengcheng.xu@uci.edu}\quad
\textsuperscript{6}\href{mailto:yu.sun@alumni.upenn.edu}{yu.sun@alumni.upenn.edu}\\
\normalsize\textsuperscript{7}\href{mailto:songchen1@xiaomi.com}{songchen1@xiaomi.com}\quad
\textsuperscript{*}Equal contribution.\quad\textsuperscript{$\dagger$}Corresponding author: Xinyu Guan.
\end{tabular}}

\begin{document}
\ninept
\maketitle
\suppressfloats[t] 
\begin{abstract}
Event-informed forecasting requires translating reports and historical responses into changes to a numerical forecast. We propose TIMBRE (Temporal Integration of Memory-Based Responses and Evidence), which combines source-aware representation, state-conditioned response transfer, and reliability-guided fusion before a frozen forecast head. A separate readout adjusts interval widths while preserving the median. In a single-seed, one-epoch development study of 13 tasks, TIMBRE improves MAE over ordinary fusion on eight tasks but over native Chronos-2 on only two. Disabling response transfer in the trained model reduces BTC and AULF MAE by 47.04\% and 6.81\%, respectively. These findings identify sensitivity to learned response transfer rather than a general forecasting advantage. Missing development-set scores and the absence of retrained ablations limit attribution to individual evidence mechanisms.
\par\noindent Code will appear at \projectlink.
\end{abstract}

\section{Introduction}
\begin{figure}[t]
\centering
\includegraphics[width=\columnwidth]{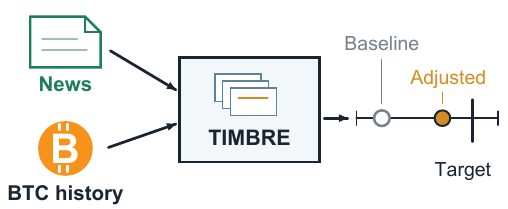}
\caption{Motivating BTC example. News and numerical history jointly inform a forecast adjustment. Target and forecast positions are schematic, not measured results.}
\label{fig:intro_example}
\end{figure}

External events may change a series before their effects appear in numerical history (Fig.~\ref{fig:intro_example}). Time-MMD pairs reports and series~\cite{timemmd_2024}; event-analysis systems structure news for forecasting~\cite{newsforecast_2024}. The question is how this evidence should modify each forecast lead.

Similar events can have different response amplitudes or delays under different numerical conditions. Directional advice needs an explicit reference, and report--score agreement may reflect shared evidence. Representations must therefore retain response shape, reference forecasts, and provenance.

\begin{figure*}[t]
\centering
\includegraphics[width=\textwidth]{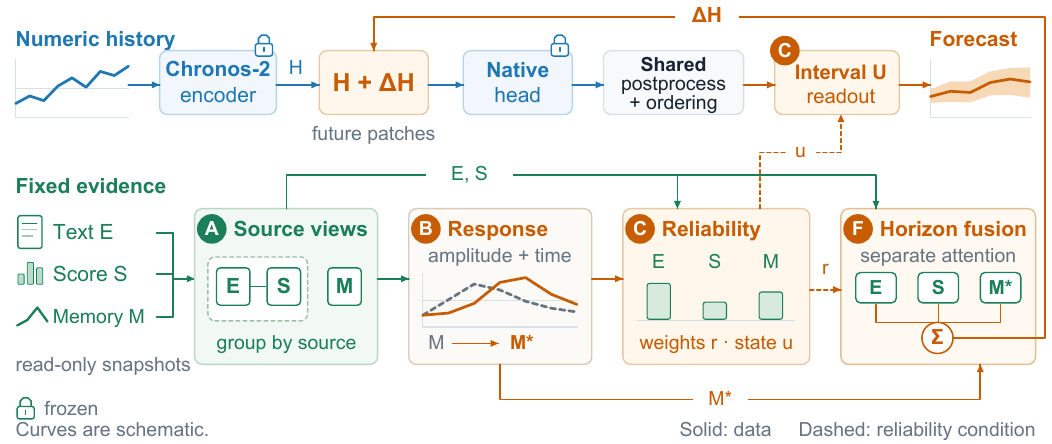}
\caption{TIMBRE's numerical and event-memory paths. Source relations (A), response transfer (B), and reliability (C) inform horizon-aware fusion (F), updating future patch states before the native head. C controls median-preserving interval readout after shared postprocessing. Numerical conditioning of B/C/F is omitted; curves and weights are schematic.}
\label{fig:architecture}
\end{figure*}

TIMBRE couples empirical historical responses and forecast-relative LLM advice with learned, state-conditioned response transfer. Source relations and jointly trained advice heads guide reliability-weighted fusion before a frozen forecast head; a separate readout adjusts interval widths (Fig.~\ref{fig:architecture}).

Component-level studies motivate attribution~\cite{llmuseful_2024}. We compare six methods and three fixed-weight interventions, reporting gains over simpler controls and degradation relative to the numerical backbone.

\par\medskip
\begingroup
\setlength{\fboxsep}{5pt}
\setlength{\fboxrule}{0.45pt}
\noindent\fbox{%
\begin{minipage}{\dimexpr\columnwidth-2\fboxsep-2\fboxrule\relax}
\setlength{\parindent}{0pt}
\setlength{\parskip}{3pt}
\textbf{Contributions}\par
\textbf{1. Adaptive event memory.} Source-aware organization and bounded amplitude/time transformations adapt retrieved response curves to the numerical state.\par
\textbf{2. Forecast-relative LLM scoring.} Source-linked horizon advice gives direction and coarse strength against a frozen reference, distinguishing abstention from neutrality.\par
\textbf{3. Joint training.} Point, quantile, and target-aligned advice supervision jointly train fusion and median-preserving intervals; pretrained models and evidence stay fixed.
\end{minipage}%
}
\endgroup
\par\medskip

\section{Related Work}
DoubleCast introduces numerical and textual decoder attention~\cite{doublecast_2026}. PIR combines instance-level error estimates with historical references~\cite{pir_2025}; PostTime learns text-conditioned forecast revision~\cite{posttime_2026}. REST-TS supervises text-specific heads with EMA numerical-predictor residuals~\cite{restts_2026}. TIMBRE instead learns state-dependent transfer and fusion of fixed scores and response curves before a frozen head.

LLMTime continues serialized values~\cite{llmtime_2023}; One Fits All adapts pretrained Transformers~\cite{onefitsall_2023}. Time-LLM maps patches to text prototypes~\cite{timellm_2024}; TimeCMA aligns numerical features with cached history-derived prompt embeddings~\cite{timecma_2025}. Our scoring interprets external reports against a numerical reference.

Numerical models use basis expansion~\cite{nbeats_2020}, linear trend/residual projections~\cite{dlinear_2023}, channel-independent patches~\cite{patchtst_2023}, or variate tokens~\cite{itransformer_2024}. Pretraining approaches include tokenized values in Chronos~\cite{chronos_2024}, patch autoregression in TimesFM~\cite{timesfm_2024}, masked reconstruction in MOMENT~\cite{moment_2024}, mixture-distribution prediction in Moirai~\cite{moirai_2024}, and lag-conditioned decoding in Lag-Llama~\cite{lagllama_2024}. Our frozen reference is Chronos-2, supporting multivariate and covariate-informed prediction~\cite{chronos2_2025}.

\section{Method}
\subsection{Fixed inputs and source organization}
At origin $t$, TIMBRE combines numerical history $X_{\leq t}$, frozen text features $E$, scores $S$, and a retrieved response $M(h)$ at physical lead $h$. Views retain source, reference, and support metadata, masking unavailable evidence. Frozen-encoder future states $\Hh$ receive a learned update before the frozen head:
\begin{equation}
 \Hh^+=\Hh+\Delta\Hh,\qquad \widehat Q^{\rm raw}=g_{\rm frozen}(\Hh^+).
 \label{eq:fusion}
\end{equation}
Module A uses relation-aware attention~\cite{relation_attention_2018} to distinguish overlapping, disjoint, and unknown sources. Canonicalization merges only exact text copies with identical authenticated sources, features, masks, and availability clocks; other views remain separate. Removing A disables both operations, retaining metadata.

\subsection{Historical response transfer}
Module B predicts bounded amplitude and timing changes from numerical state, organized evidence, and lead:
\begin{equation}
 \widetilde M(h)=a_h\operatorname{Interp}(M,\alpha h+\beta).
 \label{eq:transfer}
\end{equation}
Unlike soft-DTW's alignment loss~\cite{softdtw_2017}, this transforms a fixed response. Bounds are $a_h\in[0,2]$, $\alpha\in[0.5,2]$, and $|\beta|\leq0.25h_{\max}$, where $h_{\max}$ is the largest supported lead. Identity initialization sets $a_h=\alpha=1$, $\beta=0$. Interpolation preserves unsupported and out-of-range masks. Dispersion, counts, and references follow the time map; only responses are amplitude-scaled. Retrieval and bank contents stay fixed. Removing B restores identity transfer.

\subsection{Reliability-guided fusion}
Module C aligns score and memory advice with the current target:
\begin{equation}
 v_j(h)=b(h)+s\,d_j(h),\qquad j\in\{S,M\},
 \label{eq:advice}
\end{equation}
where $b$ is the native median and $s$ the available-history scale. Heads predict normalized residuals from numerical and evidence features, references, and support. They share relation-encoded evidence and are not independent experts. Their discrepancy, provenance, and memory statistics condition masked-softmax weights $r_E,r_S,r_M$ and interval state $u$.

Future-state and physical-lead queries cross-attend to each view~\cite{attention_2017}; reliability weights combine the reads. Updates are averaged within native patches and projected to $\Delta\Hh$, with a zero-initialized projection. Shared quantile ordering precedes the interval readout. For ordered quantiles $b_\tau$ and median $m$, it applies
\begin{equation}
 Q_\tau=m+c_\tau(b_\tau-m),\quad c_{0.5}=1,
 \label{eq:interval}
\end{equation}
using separate $c_-,c_+\in[0.5,2]$ below and above the median, initialized to one. This preserves the median. Removing C replaces reliability weights with equal available-view weights and disables this readout.

\subsection{Learning}
Only the new modules are learned, using history-scale-normalized median MAE, mean quantile pinball loss~\cite{quantiles_1978}, and supported-advice MAE:
\begin{equation}
 \calL=\calL_{\rm median}+\calL_{\rm pinball}+0.1\calL_{\rm advice}.
 \label{eq:loss}
\end{equation}
Losses average valid positions within each window, then windows. Pretrained components and upstream evidence remain fixed; targets enter losses and evaluation only, with no forecast feedback upstream.

\section{Experimental Setup}
\subsection{Data and chronological evaluation}
\begin{table}[t]
\centering
\caption{Prepared development data. FIT counts follow boundary purging; $H$ is the output-grid length. $M^+$ counts TUNE windows with at least one supported memory point, not full-horizon coverage. NHC uses a masked grid.}
\label{tab:data}
\setlength{\tabcolsep}{3pt}
\begin{tabular}{@{}lrrrr@{}}
\toprule
Data family & FIT & TUNE & $H$ & $M^+$\\
\midrule
BTC & 1,999 & 128 & 6 & 128\\
AULF & 2,805 & 705 & 48 & 705\\
NHC & 4,999 & 807 & 40 & 807\\
Time-MMD (10) & 3,990 & 599 & 6 & 588\\
\midrule
Total & 13,793 & 2,239 & -- & 2,228\\
\bottomrule
\end{tabular}
\end{table}

We evaluate BTC and Australian electricity load forecasting (AULF) as primary tasks, with NHC hurricane forecasting and ten Time-MMD domains as secondary tasks. BTC and AULF use From News to Forecast data~\cite{newsforecast_2024}; Time-MMD supplies domain-specific reports and series~\cite{timemmd_2024}. BTC predicts six daily steps and AULF 48 half-hourly steps. NHC uses a masked three-hour grid; Time-MMD retains each domain's native cadence. CAF scenarios are excluded from this comparison.

Table~\ref{tab:data} summarizes the prepared panel after removing 11 FIT windows whose final targets touch their first TUNE origin. Windows overlap and are not independent events. Missing evidence is masked. Scores average valid same-reference votes, falling back to an aligned legacy score; vote fraction is not calibrated confidence. TUNE has no supported scores, precluding an inference-time score-benefit claim. Source availability uses recorded proxy clocks; publication-time validity and score semantics are not certified, and foundation-model pretraining overlap is unaudited. All reported outcomes are development results; FINAL\_TEST remains unopened.

\subsection{Implementation and baselines}
The frozen Chronos-2 encoder and head~\cite{chronos2_2025} are augmented with 462,278 trainable parameters. Text uses frozen 1,024-dimensional token features, capped at 256 tokens per window with source-balanced quotas. Width is 64, with four attention heads and a 16-dimensional interval condition. Memory retrieval and upstream evidence are fixed. Checkpoints will be released at \weightlink.

Six methods share the TUNE panel: history mean, persistence, seasonal naive, native Chronos-2, ordinary fusion, and TIMBRE. Ordinary fusion uses identical evidence and the same insertion point, omitting A/B/C and the adaptive interval readout. Full and ordinary runs share initialization, training order, and native predictions. History mean and seasonal naive were added after full-model results were seen and are exploratory controls. Seasonal naive uses the latest matching daily phase for AULF, weekly phase for BTC, and cadence-specific annual or weekly phases for Time-MMD; NHC and unmatched phases fall back to persistence. Low-rank numerical adaptation~\cite{lora_2022} and a matched DoubleCast reproduction were not run.

Both trained methods use seed 17 and one shuffled pass without replacement: 432 AdamW~\cite{adamw_2019} updates with batch size 32, peak learning rate $10^{-4}$, weight decay 0.01, gradient clipping at 1, and cosine decay after 22 warmup updates. The last checkpoint is evaluated once; there is no hyperparameter search or TUNE-based checkpoint selection. Runs use CPU float32 with four threads. Training plus TUNE evaluation takes 34.95 minutes for TIMBRE and 33.96 for ordinary fusion, excluding upstream preparation and verification.

\subsection{Metrics and comparisons}
MAE averages valid targets within each window, then windows within an entity or storm, then these groups within a task. Table~\ref{tab:main} retains separate physical units. Pinball loss~\cite{gneiting2007strictly}, weighted interval score (WIS)~\cite{wis_2021}, and coverage use the same symmetric 21-quantile grid; on this grid WIS equals twice mean pinball loss. Simple point baselines are not assessed as calibrated distributions. The primary contrast is TIMBRE versus ordinary fusion on BTC/AULF; percentage changes average the two task-wise ratios equally. One seed supplies no across-seed standard deviation. We make no significance claims or multiplicity-controlled comparisons~\cite{holm_1979}.

\section{Results}
\subsection{Main comparison}
\begin{table*}[t]
\centering
\caption{MAE on the common TUNE panel; lower is better. BTC is in USD, AULF in MW, and NHC in knots. Time-MMD rows retain source units; Security values are divided by $10^6$ for display. $n$ counts windows. Bold highlights TIMBRE; underlining marks the lowest value among six methods, using unrounded results. History mean and seasonal naive are exploratory controls added after the trained results were seen.}
\label{tab:main}
\setlength{\tabcolsep}{5pt}
\begin{tabular}{@{}lrrrrrrr@{}}
\toprule
Task & $n$ & History mean & Persistence & Seasonal & Chronos-2 & Ordinary & \textbf{TIMBRE}\\
\midrule
AULF & 705 & 473.04 & 469.12 & 270.08 & \underline{245.52} & 250.31 & \textbf{268.80}\\
BTC & 128 & 1682.16 & \underline{311.43} & 443.23 & 322.75 & 428.72 & \textbf{649.24}\\
NHC & 807 & 14.121 & \underline{12.387} & \underline{12.387} & 13.184 & 13.569 & \textbf{14.447}\\
\midrule
\multicolumn{8}{@{}l}{Time-MMD}\\
Agriculture & 64 & 9.508 & \underline{3.240} & 5.952 & 3.938 & 4.003 & \textbf{3.999}\\
Climate & 64 & 11.005 & 4.089 & 16.006 & \underline{4.033} & 4.240 & \textbf{4.187}\\
Economy & 54 & 6204.94 & 7772.14 & 6216.22 & \underline{5247.81} & 5319.41 & \textbf{5509.29}\\
Energy & 64 & 0.31055 & \underline{0.08320} & 0.31593 & 0.08432 & 0.08909 & \textbf{0.08650}\\
Environment & 64 & 17.895 & 21.607 & 21.990 & \underline{14.683} & 14.795 & \textbf{14.771}\\
Health\_AFR & 64 & 0.05519 & \underline{0.03691} & 0.06174 & 0.03847 & 0.03872 & \textbf{0.03843}\\
Health\_US & 64 & 1.396 & 0.72790 & 0.59670 & \underline{0.40950} & 0.41051 & \textbf{0.41049}\\
Security ($10^6$) & 33 & \underline{614.39} & 668.66 & 755.19 & 673.57 & 660.06 & \textbf{642.94}\\
SocialGood & 64 & 1.523 & 0.42370 & 0.69505 & \underline{0.32569} & 0.33808 & \textbf{0.33151}\\
Traffic & 64 & 14265.28 & 18724.37 & 12668.48 & \underline{12157.83} & 12565.74 & \textbf{12847.98}\\
\bottomrule
\end{tabular}
\end{table*}

Table~\ref{tab:main} shows that TIMBRE has lower MAE than history mean on 11/13 tasks, seasonal naive on 10/13, persistence on 7/13, and ordinary fusion on 8/13. However, it improves on native Chronos-2 only for Health\_AFR and Security, by 0.11\% and 4.55\%. The primary comparison is unfavorable: BTC and AULF MAE increase by 51.44\% and 7.39\% over ordinary fusion, an equal-task mean increase of 29.41\%. Thus, wins against simpler controls do not establish additional value over the frozen numerical model.

\begin{table}[t]
\centering
\caption{Probabilistic TUNE results. PB is mean pinball loss (lower is better); C90 is coverage of the nominal 90\% interval. WIS is $2\times$PB on the shared grid.}
\label{tab:probabilistic}
\setlength{\tabcolsep}{3pt}
\begin{tabular}{@{}lrrrr@{}}
\toprule
 & \multicolumn{2}{c}{BTC} & \multicolumn{2}{c}{AULF}\\
\cmidrule(lr){2-3}\cmidrule(lr){4-5}
Method & PB & C90 (\%) & PB & C90 (\%)\\
\midrule
Native Chronos-2 & 118.27 & 91.02 & 87.05 & 83.53\\
Ordinary fusion & 168.10 & 97.27 & 88.78 & 87.09\\
TIMBRE & 239.18 & 84.24 & 96.14 & 81.67\\
\bottomrule
\end{tabular}
\end{table}

Table~\ref{tab:probabilistic} gives the same ordering for BTC/AULF pinball loss and WIS: native Chronos-2 is best and TIMBRE is worst. TIMBRE's nominal 90\% coverage is 84.24\% on BTC and 81.67\% on AULF. A trainable interval readout therefore does not, by itself, ensure calibrated or more accurate probabilistic forecasts under this training budget.

\subsection{Ablation}
\begin{table}[t]
\centering
\caption{Ablation by inference-time removal with the same trained TIMBRE weights, seed 17. All 833 BTC/AULF windows are retained; zero training updates. These are fixed-weight diagnostics, not retrained ablations.}
\label{tab:ablation}
\setlength{\tabcolsep}{3pt}
\begin{tabular}{@{}lrrrr@{}}
\toprule
 & \multicolumn{2}{c}{BTC} & \multicolumn{2}{c}{AULF}\\
\cmidrule(lr){2-3}\cmidrule(lr){4-5}
Inference mode & MAE & PB & MAE & PB\\
\midrule
TIMBRE & 649.24 & 239.18 & 268.80 & 96.14\\
w/o A: source & 483.14 & 178.66 & 263.62 & 94.04\\
w/o B: transfer & 343.86 & 126.60 & 250.51 & 88.79\\
w/o C: reliability & 731.43 & 268.70 & 272.68 & 97.65\\
\bottomrule
\end{tabular}
\end{table}

Table~\ref{tab:ablation} reports inference-time removal of each mechanism from the same trained TIMBRE checkpoint on 833 BTC/AULF windows. No optimizer updates are performed; full-model predictions exactly reproduce the saved results. Removing A disables source relations and canonicalization, removing B makes response transfer the identity, and removing C disables reliability weighting and the interval readout. C's training-time advice supervision cannot be undone by this forward-pass intervention.

Disabling B lowers BTC/AULF MAE by 47.04\%/6.81\%, while disabling A lowers it by 25.58\%/1.93\%. Disabling C instead raises errors by 12.66\%/1.44\%. Even with B disabled, MAE remains above native Chronos-2 on both tasks. These interventions identify sensitivity to learned response transfer and source organization; they do not estimate the effect of training each reduced architecture from scratch. Retrained ablations and additional seeds are outside the completed study.

\begin{figure}[t]
\centering
\includegraphics[width=\columnwidth]{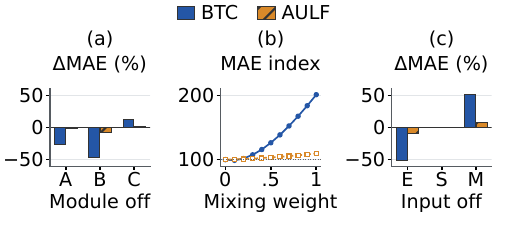}
\caption{Fixed-checkpoint diagnostics on 833 BTC/AULF TUNE windows: (a) module removal; (b) native--full output interpolation, native MAE = 100; (c) input-content masks retaining availability metadata. MAE changes are relative to FULL. Seed 17; no retraining, setting selection, or across-seed error bars.}
\label{fig:quick_diagnostics}
\end{figure}

\begin{samepage}
\noindent\textbf{Sensitivity and input controls.}
Figure~\ref{fig:quick_diagnostics} combines module removal with a fixed output-strength sweep and input-content masks. The sweep evaluates $Q_\lambda=Q_{\rm native}+\lambda(Q_{\rm full}-Q_{\rm native})$ at eleven predeclared values from zero to one, without selecting a setting. It is post-hoc output blending, not a retrained internal hyperparameter. Masking memory increases BTC/AULF MAE by 51.90\%/8.24\%, whereas masking text reduces it; masking scores leaves predictions unchanged. These masks preserve reference-availability metadata. Together, the diagnostics distinguish dependence on memory from the detrimental learned transfer in this checkpoint; they do not establish retrained component contributions.

\par\end{samepage}

\section{Discussion and Conclusion}
TIMBRE connects fixed event evidence to frozen numerical forecasts through bounded response transfer and reliability-guided fusion. In the single-seed, one-epoch development study, it improves MAE over ordinary fusion on 8/13 tasks but over native Chronos-2 on only 2/13. Disabling learned response transfer reduces BTC/AULF MAE by 47.04\%/6.81\%, identifying transfer sensitivity rather than a general accuracy gain.

Absent TUNE scores, proxy clocks, unaudited pretraining overlap, and an unopened final test limit attribution and generalization; historical responses do not establish causality. Future work will study conservative transfer and optimization, retrained ablations, and multi-seed evaluation with verified evidence availability and held-out tests.

\section{Acknowledgments}
OpenAI Codex assisted drafting and revising Sections~1--6 and writing figure-generation code.

\twocolumn
\makeatletter
\global\@colroom\@colht
\global\vsize\@colht
\makeatother
\bibliographystyle{IEEEbib}
\bibliography{references}
\end{document}